\documentclass[twoside,11pt]{article}

\usepackage{blindtext}
\usepackage{bm}
\usepackage{amsmath}
\usepackage{jmlr2e}

\usepackage{lastpage}
\ShortHeadings{bnRep: A Repository of Bayesian Networks}{Manuele Leonelli}
\firstpageno{1}

\begin{document}

\title{bnRep: A Repository of Bayesian Networks from the Academic Literature}

\author{\name Manuele Leonelli\email manuele.leonelli@ie.edu \\
       \addr School of Science and Technology\\
       IE University\\
       Madrid, Spain}

\editor{}

\maketitle

\begin{abstract}
Bayesian networks (BNs) are widely used for modeling complex systems with uncertainty, yet repositories of pre-built BNs remain limited. This paper introduces \texttt{bnRep}, an open-source R package offering a comprehensive collection of documented BNs, facilitating benchmarking, replicability, and education. With over 200 networks from academic publications, \texttt{bnRep} integrates seamlessly with \texttt{bnlearn} and other R packages, providing users with interactive tools for network exploration.
\end{abstract}


\section{Introduction}
Bayesian networks (BNs) \citep{pearl1988probabilistic} are a powerful machine learning model widely used in practice due to their ability to model complex systems with uncertainty and interpretability. Their widespread application spans diverse fields such as engineering reliability \citep{kabir2019applications}, environmental sciences \citep{kaikkonen2021bayesian}, healthcare \citep{kyrimi2021comprehensive}, and supply chain management \citep{hosseini2020bayesian}, among others. They are implemented in various software tools, including R packages (e.g. \texttt{bnlearn} \citealp{scutari2010}; \texttt{gRain} \citealp{hojsgaard2012graphical}), Python libraries (e.g. \texttt{pgmpy} \citealp{ankan2015pgmpy}; \texttt{pyAgrum} \citealp{ducamp2020agrum}), and standalone software (e.g. AgenaRisk, BayesiaLab, BayesServer, Hugin Expert, GeNie Modeler, and Netica)\footnote{Available at \url{https://www.agena.ai}, \url{https://www.bayesia.com}, \url{https://www.bayesserver.com}, \url{https://www.hugin.com}, \url{https://www.bayesfusion.com/genie/}, and \url{https://www.norsys.com/netica.html}, respectively.}, making them accessible to a wide range of users and applications. Despite their widespread application, BNs remain an active research area with ongoing advancements in learning algorithms \citep[e.g.][]{kuipers2022efficient,liu2022greedy}, inference techniques \citep[e.g.][]{lin2020improved}, and new types of networks \citep[e.g.][]{atienza2022semiparametric}.

Key reasons for the success of BNs include their explainability, flexibility, and status as the gold standard for causal modeling. Unlike black-box models, BNs offer an intuitive, transparent structure, aligning with trends in eXplainable AI (XAI) \citep{rudin2019}. They provide insights into cause-and-effect relationships rather than just correlations \citep{Peters2017}. BNs are also accessible to a broad audience, not requiring deep expertise in computer science or mathematics \citep{kelly2013selecting,moe2021increased}. Their modular design integrates data, expert knowledge, and model outputs \citep{leonelli2020coherent,marcot2019advances} and supports scenario and ``what-if" analyses for simulating interventions and predicting outcomes \citep[e.g.][]{pitchforth2013proposed}.

Despite the widespread use, flexibility, and numerous advantages of BNs, repositories of pre-built and documented BNs remain limited compared to other areas of machine learning. In fields like deep learning, there are extensive repositories such as TensorFlow Hub and PyTorch Hub that provide access to a wide range of pre-trained models\footnote{Available at \url{https://www.tensorflow.org/hub}and \url{https://pytorch.org/hub/}, respectively.}. Similarly, \texttt{scikit-learn} offers datasets and model repositories for classic machine learning algorithms, facilitating replication, benchmarking, and educational purposes. In contrast, repositories of BNs are relatively scarce, with some notable exceptions like the Bayesian Network Repository (including 31 BNs), the BNMA BN repository (including 75 BNs), and the BayesFusion Interactive Model Repository (including 46 BNs).\footnote{Available at \url{https://www.bnlearn.com/bnrepository/}, \url{https://bnma.co/bnrepo/}, and \url{https://repo.bayesfusion.com/}, respectively. } However, these collections are often smaller, provide less detailed documentation, and are not consistently updated to reflect recent advancements in BN research.

This gap motivated the creation of \texttt{bnRep}, an open-source R package that offers a comprehensive collection of documented BNs, enabling users to explore, compare, and apply them across various domains. The repository includes over 200 BNs from more than 150 different academic publications, each accompanied by detailed documentation. In the following sections, we provide a brief overview of BNs and present the key features and purposes of the \texttt{bnRep} package, highlighting why R was chosen as the development platform and how the package serves both research and practical needs.

\section{Bayesian networks}

While a full account of Bayesian networks can be found in several monographs \citep[e.g.][]{koller2009}, a brief overview is provided here to highlight the key concepts.

A BN gives a graphical representation of the relationship between a vector of variables of interest $\bm{Y}=(Y_1,\dots,Y_p)$ using a directed acyclic graph (DAG) $G$ and a factorization of the overall probability distribution $P(\bm{Y})$ in terms of simpler conditional distributions $P(Y_i\,|\, \bm{Y}_{\Pi_i})$, where $\bm{Y}_{\Pi_i}$ denotes the parents of $Y_i$ in $G$. More formally, the overall factorization of the probability distribution induced by the BN can be written as:
\begin{equation}
\label{eq:eq}
P_G(\bm{Y})=\prod_{i=1}^p P(Y_i\, |\,\bm{Y}_{\Pi_i}).
\end{equation}
This factorization has several advantages: it reduces the number of parameters to estimate, allows for easier expert elicitation of local relationships, and makes better use of available data by focusing only on the relevant variables and their parents.

The academic literature has mostly focused on three types of distributional assumptions depending on the nature of the variables: discrete, Gaussian, and conditional linear Gaussian
\citep[see e.g.][]{bodewes2021learning}. \textit{Discrete BNs} \citep{heckerman1995learning} are such that $\bm{Y}$ is a Multinomial random variable and the local distributions are defined as
\begin{equation}
\label{eq:1}
Y_i\,|\,\bm{Y}_{\Pi_i}\sim \text{Multi}(\theta_{ijk}), \mbox{ where } \theta_{ijk} = P(Y_i=j\,|\,\bm{Y}_{\Pi_i}=k).
\end{equation}
The parameters $\theta_{ijk}$ are usually reported in conditional probability tables, reporting the conditional probabilities for each parents' configuration.

\textit{Gaussian BNs} \citep{geiger1994learning} assume $\bm{Y}$ follows a multivariate Normal distribution and the local distributions are defined as linear regressions over the parents:
\begin{equation}
\label{eq:2}
Y_i\,|\, \bm{Y}_{\Pi_i} \sim \mathcal{N}\left(\beta_{i0} +\sum_{j\in\Pi_i}\beta_{ij}y_j, \sigma^2_i\right),
\end{equation}
where the $\beta$'s are the regression parameters and $\sigma^2_i>0$.

\textit{Conditional Linear Gaussian BNs} \citep{heckerman1995learning1} combine discrete and continuous random variables. Discrete $Y_i$'s can only have discrete parents and their distribution is as in Equation (\ref{eq:1}). Continuous $Y_i$'s can have both discrete and continuous parents, and their distribution is a mixture of Gaussian distributions (Equation \ref{eq:2}), one for each discrete parent configuration.

BNs can be constructed in various ways: they may be learned from data using algorithms that infer both the structure (DAG) and the conditional probabilities \citep{kitson2023survey,scutari2019learns}, elicited from expert knowledge \citep{barons2022balancing,nyberg2022bard}, or developed through a combination of the two. In practice, many models combine data-driven learning for the probabilistic parameters with expert-elicited structures or relationships, allowing for more accurate and interpretable representations of complex systems \citep{constantinou2023impact}.

\section{The bnRep package}

\subsection{Why R?}
R was chosen as the platform for \texttt{bnRep} due to its robust ecosystem for statistical modeling and seamless integration with existing tools for BNs. The \texttt{bnlearn} package \citep{scutari2010}, one of the most widely used packages for learning and inference in BNs, provides the core class objects for \texttt{bnRep}, directly supporting the three main types of BNs previously discussed. Additionally, \texttt{bnlearn} includes various exporting functions (such as \texttt{write.bif} and \texttt{write.net}), enabling users to easily transfer models between R and both Python and basically all standalone BN software. This ensures that models from \texttt{bnRep} can be utilized across different platforms and software environments, making the repository versatile for diverse applications and users' preferences.

The choice of R ensures that \texttt{bnRep} benefits from the CRAN ecosystem, which enforces clear documentation standards and regular package updates. This guarantees that \texttt{bnRep} is user-friendly, thoroughly documented, and accessible to a wide range of users. Moreover, \texttt{bnRep} is compatible with other R packages such as \texttt{gRain} \citep{hojsgaard2012graphical} and \texttt{BayesNetBP} \citep{yu2020bayesnetbp} for inference, \texttt{bnmonitor} \citep{leonelli2023sensitivity} for model diagnostics, and \texttt{qgraph} \citep{epskamp2012qgraph} for visualization, allowing for extended functionality within the R environment.

\subsection{Purposes of bnRep}

With growing interest in comparing the performance of structural learning algorithms \citep{constantinou2021large,scutari2019learns}, \texttt{bnRep} plays a central role in providing a benchmarked repository of models that enhances replicability and promotes collaboration in BN research. Researchers can test new algorithms against established models, ensuring consistent evaluation across studies. Additionally, the package serves as an educational tool, offering a diverse collection of models that help students and practitioners explore and learn about BNs across various domains \citep{de2015inference,jon}. By sharing knowledge through an accessible, well-documented, and up-to-date repository, \texttt{bnRep} fosters both innovation and cross-disciplinary applications of BNs.

\subsection{bnRep at a glance}

The \texttt{bnRep} package currently includes 214 BNs, with the number of networks constantly evolving as new models are added. Most networks come from papers published from 2020 onwards, and all are stored as \texttt{bn.fit} objects from the \texttt{bnlearn} package, which supports three types of BNs: discrete, Gaussian, and conditional linear Gaussian. These are the most widely used types of BNs in research and applications, making \texttt{bnlearn}'s support for these models crucial for \texttt{bnRep}'s extensive and versatile repository.

The package also includes the \texttt{bnRep\_summary} dataframe, which provides detailed information about each network. This includes the type of network, summaries of the DAG structure (form, number of nodes, edges, etc.), how the probabilities and DAG were defined (from data, expert knowledge, etc.), and the area of application (e.g., environmental science, engineering, medicine). Summaries of the repository and the networks' characteristics, along with visualizations, are available on the \texttt{bnRep} GitHub page (\url{https://github.com/manueleleonelli/bnRep}).

To facilitate interactive exploration, \texttt{bnRep} includes a Shiny app, accessible via the \texttt{bnRep\_app()} function. The app allows users to filter and explore the networks' database. It is freely available online at \url{https://manueleleonelli.shinyapps.io/bnRep/}, providing a user-friendly interface to access the repository without needing to install the R package.

\section{Conclusions and future directions}

The \texttt{bnRep} package provides a comprehensive, well-documented repository of BNs, making it a valuable resource for researchers and practitioners. By facilitating benchmarking, replicability, and education, \texttt{bnRep} supports the advancement of BN research and cross-field collaboration. Its integration with \texttt{bnlearn} ensures compatibility with widely-used tools, while the Shiny app enhances accessibility for users.

Future updates will expand the repository to include new types of networks, such as copula \citep{hanea2015non}, additive \citep{kratzer2023additive}, hybrid \citep[e.g.][]{perez2016parameter}, dynamic \citep[e.g.][]{shiguihara2021dynamic}, and continuous time BNs \citep{nodelman2002continuous}. The project’s open-source nature encourages contributions through GitHub, fostering community-driven growth. Additionally, benchmarking functions will be introduced, promoting the development of consistent evaluation methods in the BN field. In conclusion, \texttt{bnRep} not only addresses the current lack of documented BNs but also lays the foundation for establishing benchmarks and fostering innovation in BN research.

% Acknowledgements and Disclosure of Funding should go at the end, before appendices and references


\vskip 0.2in
\bibliography{sample}

\end{document}